\documentclass{article}
\pdfoutput=1
\usepackage{ijcai17}

\usepackage{times}

\usepackage{mybase}
\usepackage{subcaption} % replaces subfigure
\usepackage[capitalise]{cleveref} % http://ctan.sharelatex.com/tex-archive/macros/latex/contrib/cleveref/cleveref.pdf

\usepackage{ifthen}

\newcommand{\inp}{x} % inputs
\newcommand{\outp}{y} % outputs
\newcommand{\allinp}{\mathcal{X}} % all inputs
\newcommand{\alloutp}{\mathcal{Y}} % alloutputs

\newcommand{\uf}{u} % utility function
\newcommand{\allUf}{\mathcal{U}}

\newcommand{\act}{\outp} %a} % action
\newcommand{\allAct}{\alloutp}%\mathcal{A}}
\newcommand{\obs}{\inp}%e} % observation
\newcommand{\allObs}{\allinp}%\mathcal{E}}
\newcommand{\actobs}{\act\!\obs}%\mbox{\ae}}
\newcommand{\hist}{\actobs}
\newcommand{\hz}{\gamma} % discount function

\newcommand{\env}{\mu}
\newcommand{\pol}{\pi}

\newcommand{\allPol}{\mathcal{P}}

\newcommand{\poluf}{\pol_{\uf}}
\newcommand{\polufeps}{\pol_{\uf,\epsilon}}

\newcommand{\val}{\text{V}} % discounted value

\newcommand{\valenvpol}{\val^{\env, \pol}}
\newcommand{\valenvpolstaruf}{\val^{\env, \poluf^*}_{\uf}}

\newcommand{\switch}{\text{Switch}}

\newcommand{\wpolX}[1]{\weight_{\pol,#1}}

\newcommand{\mdl}{\mu}
\newcommand{\dev}{d}
\newcommand{\agt}{g}
\newcommand{\sys}{s}

\newcommand{\mixt}{M}
\newcommand{\mdev}{\mixt_\dev}
\newcommand{\magt}{\mixt_\agt}
\newcommand{\msys}{\mixt_\sys}

\newcommand{\weight}{w}
\newcommand{\wuf}{\weight_{\uf}}
\newcommand{\wdev}{\weight_\dev}

\newcommand{\weps}{\weight_\epsilon}
\newcommand{\wdevpost}[1]{\weight_{\dev, #1}}

\newcommand{\mdlset}{\mathcal{M}}
\newcommand{\devset}{\mdlset_\dev}
\newcommand{\devsetcomp}{\mdlset_{\dev}^\text{comp}}
\newcommand{\agtset}{\mdlset_\agt}

\newcommand{\cond}{|} % \mid}
\newcommand{\ocondipret}{\outp_{<t}\cond\inp_{<t}} % outs_{<t} | ins_{<t}
\newcommand{\oconditoT}{\outp_{1:T}\cond\inp_{1:T}}
\newcommand{\acondotoT}{\act_{1:T}\cond\obs_{1:T}}

\newcommand{\Tufplus}{T^{+}_{\uf}}
\newcommand{\Tufminus}{T^{-}_{\uf}}

\newcommand{\ctx}{c}
\newcommand{\allCtx}{\mathcal{C}}
\newcommand{\nctx}{C}
\newcommand{\Tctx}{T_{\ctx}}
\newcommand{\Tctxi}{T_{\ctx, i}}

\newcommand{\nact}{A}

\newcommand{\ufred}{\uf_{\text{red}}}
\newcommand{\ufgreen}{\uf_{\text{green}}}
\newcommand{\ufblue}{\uf_{\text{blue}}}
\newcommand{\ufmagenta}{\uf_{\text{magenta}}}

\newcommand{\Neps}{N_{\epsilon}}
\newcommand{\Ngoals}{N_{\text{goals}}}

\title{Agents and Devices: A Relative Definition of Agency}
\author{Laurent Orseau$^1$,
Simon McGregor McGill$^2$,
Shane Legg$^1$ \\
$^1$ DeepMind, UK\\
$^2$ University of Sussex, UK\\
}

\begin{document}
\maketitle
\date{}
%\thispagestyle{fancy}

%~ \comment{Relevant CFP: \url{http://fqxi.org/community/essay/rules}}

%~ \comment{(Olivier Pietquin) Our approach solves the zero-reward problem:
%~ If the reward is zero everywhere, all actions have the same probability
%~ and the likelihood decreases much faster.
%~ (A zero-reward may explain a random behaviour)}

%~ \todo{inconsistencies between $t$ and $T$, $<t$, $1:T$}

%~ %\warning{Most of the time we use contextual information rather than conditionals!}

%~ \todo{Shane prefers "object" than "device", as an apple is not really a device, but an object. However, a car is better described as a device than as an object?}

%~ \todo{Can we change the background knowledge in the experiments? Change the prior of the balloons, or add squares that may or may not be agents?
%~ (or something that is not a shape?)
%~ }

%~ \todo{Thm? Any behaviour can be expressed as a device or as an agent!}

\begin{abstract}
%~ Many objects can be argued to be agents, like a tree, a car, a computer,
%~ an artificial player for a computer game, a thermostat, a flyball governor for a steam engine,
%~ maybe even a stone.
%~ However, for some of these it may be
%~ How to resolve this conflict?
%Following up on Dennett's intentional stance and the work of others,
%and similarly to Russell and Norvig's hierarchy of agent designs,
%we propose a formalization of agency
%for a given system
%by contrasting two views:
%We consider that any system can be \emph{subjectively} described both as an agent
%(with goals, and maybe beliefs)
%and as a ``device'' that implements a more straightforward input-output mapping.
%This contrast allows us to ``inverse'' these two definitions
%using Bayes theorem
%to output a probability that the considered system
%fits one view better than the other.
%We validate our probabilistic (mostly Bayesian) approach on a toy grid-world domain by considering a variety of trajectories.

%\simon{

According to Dennett, the same system may be described using a `physical'
(mechanical) explanatory stance, or using an `intentional' (belief- and
goal-based) explanatory stance. Humans tend to find the physical stance more
helpful for certain systems, such as planets orbiting a star, and the
intentional stance for others, such as living animals. We define a formal
counterpart of physical and intentional stances within computational theory: a
description of a system as either a \textit{device}, or an \textit{agent}, with
the key difference being that `devices' are directly described in terms of an
input-output mapping, while `agents' are described in terms of the function
they optimise. Bayes' rule can then be applied to calculate the subjective
probability of a system being a device or an agent, based only on its
behaviour. We illustrate this using the trajectories of an object in a toy
grid-world domain.

%}
%~ In \citet{russell2003aima} terminology, we want to contrast simple reflex and model-based reflex ``agents'' (in this paper we call them \emph{devices})
%~ with goal based and utility based agents.
%~ \comment{Though we don't want to push the analogy too far!}

%~ This work originated from a group discussion with the participants of the SAB 2016 workshop on mathematical and philosophical conceptions of agency.
%\vspace*{1ex}
% tiny top-level table of contents
%\setcounter{tocdepth}{1}
%{\parskip=-2.7ex\tableofcontents}
\end{abstract}

\section{Introduction}

Humans categorise physical systems into two important classes: agents, and non-agents (which we here call `devices'). Since both are mechanically described by physics, what is the difference? Dennett has proposed that the distinction lies in how we subjectively explain these systems, and identifies two `explanatory strategies'\footnote{We ignore a third strategy, the design stance, in this article.}: the physical stance, which \citet{dennett2009intentional} describes as ``the standard laborious method of the physical sciences, in which we use whatever we know about the laws of physics and the physical constitution of the things in question to devise our prediction'', and the intentional stance, which he describes as ``the strategy of interpreting the behavior of an entity (person, animal, artifact, whatever) by treating it as if it were a rational agent who governed its `choice' of `action' by a `consideration' of its `beliefs' and `desires.'''

\citet{baker2009action} show that, by formalising agents as rational planners in an environment, it is possible to automatically infer the intentions of a human agent from its actions using inverse reinforcement learning~\citep{russell1998learning,ng2000irl,choi2015hbirl}. %,neu2007apprenticeship,ramachandran2007birl,
 However, this does not tell us whether to categorise a system as an agent or a device in the first place; this question is observer-relative, since it depends the observer's prior knowledge \citep{chambon2011what} and how efficiently they can apply each explanatory stance. % to the system.

Instead of modelling human cognition, we consider an artificial reasoner. We propose a formalization of these ideas so as to compute, from the point of view of a mechanical observer, the subjective probability that a given system is an agent. To simplify matters, we assume a clearly identified system that takes a sequence of inputs and returns a sequence of outputs at discrete time steps.

First, we discuss a few informal examples in \cref{sec:examples}.
We give some notation and the formalism of the main idea in \cref{sec:notation}.
More details on devices and agents are given in \cref{sec:devices,sec:agents}.
We validate our proposal on a set of simple experiments in \cref{sec:experiments}, showing that some behaviours are better described as devices
rather than agents, and vice-versa,
using more specific algorithms tailored for this domain.
We also demonstrate how our model can explain how agents can change their mind and switch goals---and still be considered agents, as long as the switches are  rare---thus implementing the hypothesis of \citet{baker2009action}.

%%%%%%%%%%%%%%%%%%%%%%%%%%%%%%%%%%%%%%%%%%%%%%%%%%%%%%%%%%%%%%%%%%%%%%%%
\section{Examples}\label{sec:examples}

We informally consider three examples from \citet{dennett2009intentional}: a stone, a thermostat and a game-playing computer.
%First we consider several examples to explain our point of view.

%~
%~ Basically, the more observations you get (say, when looking at a micro level),
%~ the less of an agent it becomes, because all these interactions
%~ need to be explained.

%~ \paragraph{Searle's chinese room.}
%~ Our theory can explain why people have a strange feeling when they are explained the Chinese room argument:\todo{describe it?}
%~ Because they are explained that the internal process is very simple,
%~ suddenly it feels much more complicated to explain its behaviour in terms
%~ of intentions rather than mere mechanical transformations.
%~ The sense of agency is thus lost.

%\paragraph{A stone falling through the air.}
A stone follows a parabolic trajectory when falling. If we interpret this as ``wanting to reach the ground'', we need to explain why the trajectory is parabolic rather than some other shape; it is easier to predict the trajectory directly by using Newtonian physics.

%\paragraph{A thermostat.}
\citet{dennett2009intentional} describes the thermostat as the simplest artifact that can sustain an intentional stance. The reason it is on the knife edge is that 
it can be described either as a reactive device (``if temperature is below the command, start heating''), or as an agent (``make sure the temperature is close to the command''), using descriptions of comparable simplicity.

%However, if we dig just a little deeper into the details, we find that the rule may be similar to ``if temperature is less than $1^{\circ}$C below the command, start heating for 10mn'', which seems a bit more difficult to express in terms of goals, and can leave us with a feeling of the thermostat being more like a reactive system than an agent. However, if the thermostat was able to close the doors and the windows and behave in a much more complex way so as to always maintain the temperature to a given command, one then may be more inclined to call it an agent.

%\paragraph{An adversary.}
%Imagine that you are trying to achieve some goal, but the system in front of you
%seems to keep doing its best to prevent you from getting there.
%Each time you try a new strategy, the system finds a new way to block you.
%Such a system would likely be regarded as an agent, as its goals seems simple (``to block you'')
%whereas describing its behaviour without this goal seems complex.
%This may explain why we start being angry at and shouting at our computer as if it would change something
 %when it seems to resist all attempts at making it function normally.
 %Such an example is typical of two-player games, even if the adversary is an artificial player.

%\paragraph{A game-playing computer.}
A system may strongly invite the intentional stance even if it is entirely reactive. For example, the policy network in AlphaGo~\citep{silver2016alphago} can play go at a high level, even without using Monte-Carlo tree search. A mechanical description would be fairly complex, consisting mostly of a large list of apparently arbitrary weights, but it is very simple to express the goal ``it wants to win at the game of go''. 

\section{Notation and formalism}\label{sec:notation}

At each time step $t$, the system under consideration receives
an input or observation $\inp_t\in\allinp$ and returns an output or action $\outp_t\in\alloutp$.
We denote history pair $(\inp_t, \outp_t)$ by $\hist_t$.
These produce the sequences $\inp_{1:t}$ and $\outp_{1:t}$ of inputs and outputs
from step 1 to $t$ included, and we call the sequence $\hist_{1:t}$
an \emph{interaction history} or \emph{trajectory}.
We will also use the notation $\inp_{<t}\equiv \inp_{1:t-1}$,
and similarly for $\outp$ and $\hist$.
The sets $\allinp$ and $\alloutp$ are considered finite for simplicity.
The probability simplex over a set $S$ is denoted $\probsimplex{S}$, \ie
if $w\in \probsimplex{S}$, then $\forall i\in[1..|S|]: w_i \in[0,1]$ and $\sum_{i=1}^{|S|} w_i = 1$.
The indicator function $\indicator{test}\in{0,1}$ has value 1 if $test$ is true, 0 otherwise.

%%%%%%%%%%%%%%%%%%%%%%%%%%%%%%%%%%%%%%%%%%%%%%%%%%%%%%%%%%%%%%%%%%%%%%%%
%\paragraph{Formalism.}
In order to output a probability that a system is an agent,
we must give probabilistic definitions of both devices and agents
and then apply Bayes theorem to inverse the likelihood of an
\emph{observed} trajectory to posterior probabilities of both views of the system.
We take a Bayesian point of view: a system belongs to a set of possible systems, so we build a mixture of all such systems for both agents and devices.

\paragraph{Describing devices: Mixture $\mdev$.}
%~ The mixture of devices looks very much like Solomonoff's mixture,
%~ except that we contextualize on the inputs, similarly to \citet{Hutter2004uaibook}.
%
Let $\devset$ be a set of physical processes that can be described
as a system, \ie as an input-output device, that is, as some function $\dev:\left(\allinp\times\alloutp\right)^*\to\probsimplex\alloutp$
that outputs a probability distribution to outputs given an interaction
history of inputs $\inp_{<t}$ and outputs $\outp_{<t}$.
The set $\devset$ can be finite, countable, or uncountable, but we consider it countable here.
Then the likelihood of the sequence of outputs for a given sequence of inputs to the system, and supposing that the system is a device is
\begin{align*}
P(\ocondipret, \mdev) &= \sum_{\dev\in\devset} P(\dev)P(\ocondipret, \dev), \\
\text{that is, }\mdev(\ocondipret) &:= \sum_{\dev\in\devset} \wdev\dev(\ocondipret).
\end{align*}
$\mdev$ is thus a mixture of all these probability distribution functions,
where each such function $\dev$ is assigned a \emph{prior weight} $\wdev\in[0, 1]$
so that $\sum_{\dev\in\devset} \wdev = 1$.

Among all device descriptions in $\devset$,
at step $t$
the posterior probability $\wdevpost{t}$ of a particular device description $\dev$
is found using Bayes rule in sequence:
%\[
$
\wdevpost{t} :=
\frac{\wdev\dev(\ocondipret)}{\mdev(\ocondipret)}
$
%\]
and the conditional probability of the next output can now be written:
%\begin{align*}
$
\mdev(\outp_t\cond\outp_{<t},\inp_{<t}) = \sum_{\dev\in\devset} \wdevpost{t}\dev(\outp_t\cond\outp_{<t},\inp_{<t}).
$
%\end{align*}

\paragraph{Describing agents: Mixture $\magt$.}
Similarly to devices, we define a mixture over the set $\agtset$ of all possible agents $\agt\in\agtset$.
We will describe how to define the mixture $\magt$ and the models for the agents in Section~\ref{sec:agents}.

\paragraph{Putting it altogether: Mixture $\msys$.}
Now we can put both descriptions together in a single mixture $\msys$.
In effect, within $\msys$ we assume that any trajectory can be explained by either the mixture of agents or the mixture of devices, and nothing else.
We take an uniform prior of the two mixtures:
\begin{align*}
\msys(\ocondipret) := \half\mdev(\ocondipret) + \half\magt(\ocondipret).
\end{align*}

Using Bayes' rule, we can now compute the likelihood that a sequence of outputs $\outp_{<t}$
is generated by an agent rather than by a system.
The (subjective) probability that the device is an agent
given a trajectory
is the probability that the trajectory is generated by an agent with the environment
times the prior probability of being an agent ($\half$):
\begin{align*}
P&(\text{agt} \cond \outp_{<t}, \inp_{<t})
%=
%\frac{P(\ocondipret,\text{agt})P(\text{agt})}{P(\ocondipret)} \\
%&=\frac{P(\ocondipret,\text{agt})P(\text{agt})}{
%P(\ocondipret,\text{agt})P(\text{agt})
%+ P(\ocondipret,\text{dev})P(\text{dev})
%}
%\\
%&=
%\frac{\half\magt(\ocondipret)}{\msys(\ocondipret)}
= \frac{\magt(\ocondipret)}{\mdev(\ocondipret) + \magt(\ocondipret)} \\
P&(\text{dev} \cond \outp_{<t}, \inp_{<t})
%= 1- P(\text{agt} \cond \outp_{<t}, \inp_{<t})\\
= \frac{\mdev(\ocondipret)}{\mdev(\ocondipret) + \magt(\ocondipret)}.
\end{align*}

Furthermore, the posterior probability of a particular device $\dev\in\devset$,
\ie how well this device can explain the trajectory compared to other
devices and agents, is
\begin{align*}
P&(\dev\cond\inp_{<t}, \outp_{<t})
= P(\mdev\cond\inp_{<t}, \outp_{<t}) P(\dev\cond\inp_{<t}, \outp_{<t},\mdev) \\
&=
 \frac{\half\mdev(\ocondipret)}{\msys(\ocondipret)}
 \frac{\wdev\dev(\ocondipret)}{\mdev(\ocondipret)}
 =
 \frac{\half\wdev\dev(\ocondipret)}{\msys(\ocondipret)}
\end{align*}
and similarly for an agent $\agt$.

\subsection{Devices}\label{sec:devices}

In principle, the device mixture $\mdev$ can be any probabilistic model
that can be used to compute a likelihood of the output history;
A more Bayesian view is to consider the set of all possible models (decision trees, neural networks, etc.) within some class
and assign some prior to them.
In \cref{sec:experiments} we use a mixture of simple contextual predictive models.

To produce a complete inference algorithm, we also consider
the choice of a universal prior measures over the set of all computable devices.

\paragraph{Information theoretic choice: Algorithmic probability.}
%Ignoring computational limitations,
%an optimal choice for the device mixture is to use (a straightforward variant of) Solomonoff's mixture~\citep{solomonoff1964formal,legg2008machine}.
%It places a (subjective) prior over all (semi-)computable devices
%(possibly incorporating background knowledge),
%such that if the observed input-output trajectory can be described by \emph{any} computable
%function, Solomonoff's inference will quickly learn to predict correctly its behaviour,
%as it assigns a non-vanishing weight to this function, and discards all functions that are inconsistent with the observed sequence.
%First, we need to choose a reference Turing machine, which is equivalent to
%choosing a (Turing-complete) programming language of reference.
%In such a language, all (semi-)computable devices $\devsetcomp$ can be expressed:
%Consider a program $\dev$ that, given a sequence of inputs $\inp_{<t}$ and outputs $\outp_{<t}$,
%outputs a probability distribution%
%\footnote{which can be simply a set of real numbers that is normalized afterwards.}
 %over the next observation $\outp_t\in\alloutp$.
%On this reference machine, each device $\dev\in\devsetcomp$ is assigned a prior weight $\wdev := 2^{-\ell_\dev}$,
%where $\ell_\dev$ is the length in bits of the description of the device on the reference machine.
%Kraft's inequality~\citep{li2008introduction} ensures that $\sum_{\dev\in\devsetcomp} 2^{-\ell_\dev} \leq 1$
%if the functions are encoded in a prefix-free way.%
%
Ignoring computational limitations,
an optimal choice for the device mixture is to use (a straightforward variant of) Solomonoff's mixture~\citep{solomonoff1964formal,legg2008machine} for some particular Turing-complete reference machine. If an observed input-output trajectory can be described by \emph{any} computable
function, Solomonoff's inference will quickly learn to predict correctly its behaviour. In the programming language for our reference machine,
all (semi-)computable devices $\devsetcomp$ can be expressed:
Consider a program $\dev$ that, given a sequence of inputs $\inp_{<t}$ and outputs $\outp_{<t}$,
outputs a probability distribution
%\footnote{which can be simply a set of real numbers that is normalized afterwards.}
 over the next observation $\outp_t\in\alloutp$.
Each device $\dev\in\devsetcomp$ is assigned a prior weight $\wdev := 2^{-\ell_\dev}$,
where $\ell_\dev$ is the length in bits of the description of the device on the reference machine.
%~ Kraft's inequality~\citep{li2008introduction} ensures that $\sum_{\dev\in\devsetcomp} 2^{-\ell_\dev} \leq 1$
%~ if the functions are encoded in a prefix-free way.%
%~ \footnote{For example, programs that end with ``ENDPROG'' are prefix-free, since whatever comes after this instruction is disregarded.}
%~ The Solomonoff mixture is then defined by
%\begin{align*}
$
\mdev(\ocondipret) := \sum_{\dev\in\devsetcomp}2^{-\ell_\dev} \dev(\ocondipret).
$
%\end{align*}
Hence, if there is a computable device $\dev^*$ that correctly describes
the system's behaviour (\ie if the system's behaviour is computable),
then Solomonoff's mixture prediction will be almost as good as $\dev^*$
since at all steps $t$, $\mdev(\ocondipret) \geq \weight_{\dev^*} \dev^*(\ocondipret)$
or in logarithmic-loss or code redundancy terms
%\begin{align*}
$
\forall t:
\log \frac{1}{\mdev(\ocondipret)} - \log \frac{1}{\dev^*(\ocondipret)} \leq \log \frac{1}{\wdev}.
$
%\end{align*}
Thanks to this very strong learning property, the subjective prior bias
quickly vanishes with evidence, that is, with the length of the trajectory.
%\quest{Add comment on being semi-measure?}

\paragraph{A (somewhat) more computable choice.}
%There are two main issues with using a Solomonoff prior.
%The first one is that it is not computable, but this is not a surprise.
%The second one is more specific to the current work: In our case,
%agents can also be programmed in a programming language,
%and not considering computation time means that
%the device mixture can mostly do this ``for free'' (invariance theorem~\citep{li2008introduction}).
%To fix both problems at the same time, we can use instead the speed prior~\citep{schmidhuber2002speed,filan2016loss},
%which also takes into account the computation time required to output the sequence $\outp_{<t}$:
%it is computable (though not tractable) and the invariance theorem is weakened because there is a computation-time cost for implementing interpreters.
%
Under a Solomonoff prior (which does not consider computation time), the
invariance theorem~\citep{li2008introduction} says the prior also contains
an ``interpreter'' for all agents. The cost to describe an agent as a device
is then always bounded by the cost of the interpreter.
The speed prior~\citep{schmidhuber2002speed,filan2016loss} is a computable variant of the Solomonoff prior that takes into account the computation time required to output the sequence $\outp_{<t}$, hence greatly weakening the invariance theorem.

%There are two main issues with using a Solomonoff prior.
%The first one is that it is not computable, but this is not a surprise.
%The second one is more specific to the current work: In our case,
%agents can also be programmed in a programming language,
%and not considering computation time means that
%the device mixture can mostly do this ``for free'' (invariance theorem~\citep{li2008introduction}).
%To fix both problems at the same time, we can use instead the speed prior~\citep{schmidhuber2002speed,filan2016loss},
%which also takes into account the computation time required to output the sequence $\outp_{<t}$:
%it is computable (though not tractable) and the invariance theorem is weakened because there is a computation-time cost for implementing interpreters.

%\todo{Prior knowledge? Computation limitations? Simplifications? (\eg, Newtonian physics vs quantum physics, or even naive physics \url{https://en.wikipedia.org/wiki/Na\%C3\%AFve_physics}.}

A more observer-dependent prior could also be considered,
for example that depends on the computational limitations of the observer
and its background knowledge about the world.

%%%%%%%%%%%%%%%%%%%%%%%%%%%%%%%%%%%%%%%%%%%%%%%%%%%%%%%%%%%%%%%%%%%%%%%%
\subsection{Agents}\label{sec:agents}

To assess whether a given trajectory is agent-like,
we apply Bayesian inverse reinforcement learning~\citep{ramachandran2007birl,choi2015hbirl} %,rothkopf2011preference}
 except that we want
to output a probability rather than a reward function.
%Hence, by contrast to previous approaches, we do not seek the maximum a posteriori element.

Since the problem is inherently harder than ``forward'' RL, most previous work in IRL focuses on MDPs.
Here, since the purpose of this paper is to provide a unified and general framework,
we propose a more general formulation
using Bayesian model-based and history based environments~\citep{Hutter2004uaibook}.
The model of the environment may be imperfect and allows for the agent to learn about it through interaction
(and update its beliefs with Bayes theorem).
For agents, inputs are usually called \emph{observations} and outputs \emph{actions}.

After describing this general reinforcement learning framework,
we ``invert'' it to find the probability that an agent is
acting according to some reward function.

An \emph{environment} $\env: (\allAct\times\allObs)^*\to \probsimplex \allObs$ is a probability distribution
%%
%\footnote{A semi-measure is a measure where the sum of the probabilities may be less than 1, that is, it assigns a non-zero probability to the non-observation symbol $\noObs$. A semi-measure can be renormalized to a proper measure, but note that a renormalized semi-computable semi-measure may not be semi-computable any more.}
 over observations given the past observations and actions, with
$
\env(\obs_{t:T}|\act_{1:T}, \obs_{<t}) := \prod_{k=t}^T \env(\obs_k | \obs_{<k}, \act_{1:k}).
$
The environment can either be the known environment or an uncertain environment,
as in a mixture of potential environments, with their posteriors updated using Bayes theorem.
%\begin{align*}
%\end{align*}
%\comment{Chronological measures? The definition above would be well defined already.}

A \emph{utility function} (or reward function) $\uf: (\allAct\times\allObs)^*\to [0, 1]$ assigns an instantaneous value to the current trajectory.
%\footnote{Considering unbounded utilities leads to various problems, hence we restrict them to a bounded interval which can then be set to $[0, 1]$ without loss of generality.}
The cumulated utility of an interaction sequence is the sum
of the instantaneous utilities along that sequence.

A \emph{policy} $\pol: (\allAct\times\allObs)^*\to\probsimplex \allAct$
is a probability distribution over actions given the past, \ie
$\pol (\act_t \cond \hist_{<t})$ is how likely the agent is to take
action $\act_t$ at time $t$.
Similarly to environments, we extend the definition of a policy:
%\begin{align*}
$
\pol(\act_{t:T}|\obs_{<T}, \act_{<t}) := \prod_{k=t}^T \pol(\act_k | \act_{<k}, \obs_{<k}).
$
%\end{align*}

Now, given a particular utility function $\uf$, the value of a given policy $\pol$
in an environment $\env$ is given by:
\begin{align}\label{eq:valenvpoluf}
\valenvpol_\uf(\hist_{<t}, \act_t) &:=
\expect_{\obs_t\sim\env}
%\sum_{\obs_t} \env(\obs_t \cond \hist_{<t}, \act_t)
\left[
\uf(\hist_{1:t}) + \hz
\expect_{\act'\sim\pol}
%\sum_{\act'}\pol(\act'|\hist_{1:t})
\valenvpol_\uf(\hist_{1:t}, \act')
\right].
\end{align}
where $\hz\in[0,1)$ is the \emph{discount factor}. % ]$
This last form also allows us to consider the value of taking action $\act_t$ after some history $\hist_{<t}$, which is useful to define the policies.
In particular, we may want the agent to follow the \emph{best} policy $\pol^*_{\uf}$ that always chooses one of the actions of optimal value for a given underlying utility function $\uf$ in an environment $\env$:
\begin{align*}%\label{eq:argmaxV}
\allAct^*_t &:= \arg\max_{\act'} \valenvpolstaruf(\hist_{<t}, \act') \\
\poluf^*(\act_t|\hist_{<t}) &:= \frac{1}{|\allAct^*_t|}\indicator{\act_t\in\allAct^*_t}.
\end{align*}
%where $\indicator{test}\in\{0,1\}$ is the indicator function.

But it is more realistic to consider that the agents are only approximately
rational.
%To this end, it is common to use a use a $\beta$-softmax policy for some parameter $\beta\in(0,+\infty)$, called the temperature (\eg \citep{baker2009action,choi2015hbirl,neu2007apprenticeship}).
%\begin{align*}
%\pol_{\uf,\beta}(\act_t \cond \hist_{<t})
%&:= \frac{\exp\left(\beta\valenv_\uf(\hist_{<t}, \act)\right)}{\sum_{\act'}\exp\left(\beta\valenv_\uf(\hist_{<t}, \act')\right)}.
%\end{align*}
%If $\beta=0$, $\pol_\beta$ chooses its actions uniformly, and if $\beta=\infty$, it reverts to \cref{eq:argmaxV}.
%~ Note that we use the \emph{on-policy value} of the agent,
%~ which means the agent may be aware of its rational limitations
%~ and try to cope with it, like Sarsa~\citep{sutton1998reinforcement}.
%~ \todo{Include the parameter $\beta$ in the models of the agents, so that it
%~ can be inferred.}
%Although inference of the parameter $\beta$ is feasible (see for example~\citet{rothkopf2011preference}),
For simplicity in the remainder of this paper
we will consider $\epsilon$-greedy policies instead,
which is still one of the favourite choices in RL research (\eg \citet{mnih2015dqn}).
The policy of the $\epsilon$-greedy agent chooses an optimal
action with probability
 $1-\epsilon$:
\begin{align}\label{eq:epsgreedy}
\pol_{\uf,\epsilon}(\act_t \cond \hist_{<t})
&:=
%(1-\epsilon)\frac{1}{|\allAct^*_t|}\indicator{\act_t\in\allAct^*_t}
%+ \notag\\
%&\quad\epsilon\frac{1}{|\allAct\setminus\allAct^*_t|}
%\indicator{\act_t\notin\allAct^*_t} \notag\\
%&=
\left(\frac{(1-\epsilon)}{|\allAct^*_t|}\right)^{\indicator{\act_t\in\allAct^*_t}}
\left(\frac{\epsilon}{|\allAct\setminus\allAct^*_t|}\right)^{\indicator{\act_t\notin\allAct^*_t}}\hspace{-1cm}.\hspace{.5cm}
\end{align}
With $\epsilon=0$, the agent always selects one of the best actions, that is,
it acts rationally.%
\footnote{This definition slightly departs from the standard one
$(1-\epsilon)\frac{1}{|\allAct^*_t|}\indicator{\act_t\in\allAct^*_t}
+ \epsilon\frac{1}{|\allAct|}$
in order to allow for integrating over $\epsilon$.
%$(1-\epsilon)\frac{1}{|\allAct^*_t|}\indicator{\act_t\in\allAct^*_t}
%+ \epsilon\frac{1}{|\allAct|}$.
}
%\todo{Make $\allAct^*_t$ depend on $\uf$?}

\paragraph{Inference.}
In an environment $\env$, given a utility function $\uf$ and an exploration parameter $\epsilon$, we can compute the likelihood of the sequence of actions $\act_{<t}$ conditioned on the observations $\obs_{<t}$ simply
with $\pol_{\uf, \epsilon}(\act_{<t}|\obs_{<t})$.

Thanks to the nice form of \cref{eq:epsgreedy}, we can actually make a mixture $\poluf$ of all values for $\epsilon$ in closed form:
\begin{align*}
&\poluf(\acondotoT)
:=\int_0^1 \polufeps( \acondotoT) \weps\, \D\epsilon \\
&= \int_0^1 \prod_{k=1}^T
\left(\frac{(1-\epsilon)}{|\allAct^*_t|}\right)^{\indicator{\act_t\in\allAct^*_t}}
\left(\frac{\epsilon}{|\allAct\setminus\allAct^*_t|}\right)^{\indicator{\act_t\notin\allAct^*_t}}
%\left(\frac{(1-\epsilon)}{|\allAct^*_k|}\indicator{\act_k\in\allAct^*_k} + \frac{\epsilon}{|\allAct|-|\allAct^*_k|}\indicator{\act_k\notin\allAct^*_k}\right)
 \weps\, \D\epsilon\\
&=
\prod_{k=1}^T
\frac{1}{|\allAct^*_t|^{\indicator{\act_t\in\allAct^*_t}}\,
|\allAct\setminus\allAct^*_t|^{\indicator{\act_t\notin\allAct^*_t}}}
%\prod_{k=1}^T
%\left(\frac{\indicator{\act_k\in\allAct^*_k}}{|\allAct^*_k|} + \frac{\indicator{\act_k\notin\allAct^*_k}}{|\allAct|-|\allAct^*_k|}\right)
\int_0^1 (1-\epsilon)^{\Tufplus}\epsilon^{\Tufminus} \weps\, \D\epsilon
\end{align*}
where $\weps$ is some prior over $\epsilon$ and
 $\Tufplus := |\{t\in[1..T]:\act_t\in\allAct^*_t\}|$ is the number of times a best action is chosen w.r.t. $\uf$, and $\Tufminus := T-\Tufplus$.
The integral is the definition of the Beta function, and thus taking
$\weps:=1$ we obtain:
\begin{multline}\label{eq:eps_integrated}
\poluf(\acondotoT)
= \\
\left(\prod_{\substack{k=1 \\ \act_k\in\allAct^*_k}}^T \frac{1}{\allAct^*_k}
\right)\left(\prod_{\substack{k=1 \\ \act_k\notin\allAct^*_k}}^T\frac{1}{|\allAct|-|\allAct^*_k|}\right)
\frac{1}{T+1}\binom{T}{\Tufplus}^{-1}
\end{multline}
where $\binom{a}{b}$ is the binomial coefficient $\frac{a!}{b!(a-b)!}$.

Finally, we can now build the mixture $\magt$ over all goals:
\begin{align}\label{eq:magt}
\magt(\acondotoT) &:=
\sum_{\uf\in\allUf} \wuf \poluf(\acondotoT).
\end{align}
A simple choice for the weights $\wuf$ is $\wuf := \frac{1}{|\allUf|}$ if $\allUf$ is finite.

%~ Computing $\magt(\acondotoT)$ is efficient as long as the number of goals
%~ is small enough, but for a large set of goals it may require additional tricks
%~ that can take into account the structure of these goals; this is not the purpose of the present work so as to focus on clarity
%~ (see previous work for various ways to perform efficient inference
%~ in MDPs, \eg \citet{ng2000irl,neu2007apprenticeship,choi2015hbirl,ramachandran2007birl,rothkopf2011preference}).
%~ See previous work for efficient inference for
%~ \todo{See \citet{?} for efficient inference.}

%~ % no MAP
%~ Note that by contrast to most previous in IRL we do not seek the Maximum A Posteriori (MAP) estimate, and rely on the likelihood of the whole mixture,
%~ which is usually dominated by the MAP model, but may not always be.
%~ \comment{Already said earlier.}

\paragraph{Universal IRL.}
Similarly to devices in \cref{sec:devices}, we can also use Solomonoff's prior over the set of reward functions,
which would lead to ``inverting'' AIXI, where AIXI is the optimal Bayesian RL agent for the class
of all computable environments and reward functions~\citep{Hutter2004uaibook}.

\paragraph{With the speed prior for devices.}
In the case we use the speed prior for the devices,
one problem arises: Since the agent can use the Bellman equation
for free, if any device can be represented as an agent then
everything may look like an agent because the penalty for devices is too large.
To compensate for this, we take away something from agents,
for example we can set he prior to $2^{-2\ell(\uf)}$ instead of $2^{-\ell(\uf)}$.
%~ so as to give a
%~ which means that if we try to express a device as a agent this may
%~ double the complexity, and thus give an advantage to devices.

%%%%%%%%%%%%%%%%%%%%%%%%%%%%%%%%%%%%%%%%%%%%%%%%%%%%%%%%%%%%%%%%%%%%%%%%
\section{Experiments}\label{sec:experiments}

To test our hypothesis, we built a gridworld simulator (see for example Fig.~\ref{fig:magenta}).
The system under consideration (the yellow triangle)
can move in the 4 directions (up, down, left, right) except if there is a wall.
The red, green, blue and magenta balloons have fixed positions.
Does the system
act rationally according to one of the goals, or is its behaviour
better described as a moving device that simply reacts to its environment?
The experimenter can make the triangle follow a sequence of actions
$\act_{1:T}$.

%\begin{figure}[htbp!]
%\centering
%\includegraphics[height=4cm]{img/simulator.png}
%\caption{The system under consideration is the yellow triangle.
%The red, green, blue and magenta balloons are potential goals. Does the system
%act rationally according to one of the goals, or is its behaviour
%better described as a moving device that simply reacts to its environment?}
%\label{fig:simulator}
%\end{figure}

\subsection{Device descriptions}

For a device, we define the observation $\obs_t$ at step $t$ to be the kind of cell
(wall, empty, red, green, blue, magenta)
it is facing in the world, in the direction of its last action.

A device's behaviour is defined by a set of associations between
a context and an action, for all possible contexts;
a context is made of the current observation and the last action the agent took.
%\footnote{Because the context on the very first step is not meaningful, we simply discard the first step.}
%
An example of a device's deterministic function can be found in \cref{tab:exdev}.
\begin{table}
\centering
\caption{An example of a device that moves along the walls.}
\begin{tabular}{c|c|c|c|c|c|c|}
&\multicolumn{6}{c|}{Cell in front of the system} \\
\cline{2-7}
  &wall & empty & red & green & blue & magenta\\
Last act& &&&&&\\
\hline
$\uparrow$& $\leftarrow$ & $\uparrow$ & $\uparrow$ & $\uparrow$ & $\uparrow$ & $\uparrow$\\
$\downarrow$& $\rightarrow$ &$\downarrow$&$\downarrow$&$\downarrow$&$\downarrow$ & $\downarrow$ \\
$\leftarrow$& $\downarrow$ &$\leftarrow$&$\leftarrow$&$\leftarrow$&$\leftarrow$ & $\leftarrow$\\
$\rightarrow$& $\uparrow$ &$\rightarrow$&$\rightarrow$&$\rightarrow$&$\rightarrow$ & $\rightarrow$
\end{tabular}
\label{tab:exdev}
\end{table}
%~ \begin{table}
%~ \centering
%~ \caption{An example of a reactive device that moves along the walls.}
%~ \begin{tabular}{l|c|c|c|c|c|c|}
%~ &\multicolumn{6}{c|}{\textbf{Cell in front of the system}} \\
%~ \cline{2-7}
  %~ &wall & empty & red & green & blue & magenta\\
%~ \textbf{Last act.}& &&&&&\\
%~ \hline
%~ up& left & up & up & up & up & up\\
%~ down& right &down&down&down&down & down \\
%~ left& down &left&left&left&left & left\\
%~ right& up &right&right&right&right & right
%~ \end{tabular}
%~ \label{tab:exdev}
%~ \end{table}
%
There are $4^{24}\approx 10^{14}$ different deterministic functions describing devices.
%~ There are $4\times 6 = 24$ possible contexts in which an action must be specified,
%~ and since there are 4 possible actions per context,
%~ this means there are
%
%~ Now, for a given trajectory it may happen that the experimenter
%~ takes two different actions in the same context at two different time steps.
%~ In that case, no single deterministic device can fit such a description.
%~ Hence, a
As for agents below, we allow for $\epsilon$-deterministic devices,
\ie at each step there is a probability of $1-\epsilon$ that the device
takes the agent given by its deterministic function, and an $\epsilon$
chance that it takes a different action.

Each context is associated with a multinomial predictor.
Let $\nact := |\allAct|$ be the number of actions.
Let $\allCtx$ be the set of all mutually-exclusive contexts (only one context is active at any step),
%\footnote{In a mutually exclusive context set, at each step exactly one context is active and used for prediction.}
 and let $\allCtx_T$ be the set
of contexts that have been visited after the trajectory $\hist_{1:T}$.
Let $\Tctxi$ be the number of times action $\act = i$ has been taken
in the context $\ctx$, and let $\Tctx := \sum_{i=1}^{\nact} \Tctxi$ be the number of visits of the context $\ctx$.
An $\epsilon$-deterministic context model puts a categorical distribution
over the set of actions for each context,
where $\epsilon$ is a $\nctx$-dimension vector
of probability distributions over $\allAct$,
hence $\forall\ctx, \sum_{i=1}^{\nact} \epsilon_{\ctx,i}=1$:
%\begin{align*}
$
\mdl_{\ctx, \epsilon}(\oconditoT) %&
:=
\prod_{\ctx\in\allCtx}(\epsilon_{\ctx,i})^{\Tctxi},
$
%\end{align*}
which in the current experiments are essentially a Markov model of order 2.
We can now build a continuous mixture of all such $\epsilon$-deterministic context models:
\begin{align*}
&\mdev(\oconditoT) \\
&:= \underbrace{\int_0^1\!\!\!\!\cdots\!\!\int_0^1}_{\nact-1 \text{ times}}
\prod_{\ctx\in\allCtx}
\mdl_{\ctx,\epsilon}(\oconditoT)
p(\epsilon_{\ctx,2})\D\epsilon_{\ctx,2}\cdots p(\epsilon_{\ctx,\nact})\D\epsilon_{\ctx,\nact} \\
%~ &=
%~ \prod_{\ctx\in\allCtx}
%~ \int_0^1\!\!\!\!\cdots\!\!\int_0^1
%~ \mdl_{\ctx,\epsilon}(\oconditoT)
%~ p(\epsilon_{\ctx,2})\D\epsilon_{\ctx,2}\cdots p(\epsilon_{\ctx,\nact})\D\epsilon_{\ctx,\nact}
\end{align*}
where $\epsilon_{\ctx, 1} = 1-\sum_{i=1}^{\nact-1}\epsilon_{\ctx, i}$.
Taking a uniform prior $p(\epsilon_{\ctx,i}) := 1$ over $\epsilon$
leads to a multinomial estimator:% for each context:
\begin{align*}
\mdev(\oconditoT)
:= \prod_{\ctx\in\allCtx}(\nact-1)!\frac{\prod_{i=1}^{\nact}(\Tctxi!)}{(\Tctx+\nact-1)!}.
%\frac{\prod_{i=1}^\nact \prod_{k=0}^{\Tctxi-1}(k+\half)}{\prod_{k=0}^{\Tctx-1}(k+\half\nact)}
\end{align*}

\subsection{Agent descriptions}

%\comment{Due to the simplicity of the models, we perform full inference of the mixture.}

We consider a very small set of $\Ngoals :=4$ goals, $\allUf := \{\ufred, \ufgreen, \ufblue, \ufmagenta\}$ ---the red, green, blue, and magenta circles in \cref{fig:magenta}.

To be able to assign a probability to the actions of the trajectory,
we first need to solve the Markov Decision Process (MDP)~\citep{sutton1998reinforcement} for each goal,
using \emph{states} instead of histories,
where the state is simply a (row, column) position in the environment.
The value $\valenvpol_{\uf}$ in Eq.~\eqref{eq:valenvpoluf} is then computed
for each state-action,
with a reward of 1 for reaching the goal, and 0 everywhere else.
%~ For a given goal, the reward is 1 if the agent reaches the goal, and 0 everywhere else.
%Together with using $\epsilon$-greedy policies rather than softmax policies,
%this has the useful property that in any given state, the relative values of the actions are independent of the discount factor $\hz$; thus for each action we only need to know the minimum distance to the goal,
%which can be found using Dijkstra's shortest-path algorithm~\citep{dijkstra1959note}: in each state, either the selected action is on the shortest path from this state, or it is not.
%
%~ For $\epsilon$-greedy policies, the relative values of the actions are independent of the discount factor $\hz$ given the state; thus for each action we only need to know the minimum distance to the goal,
%~ which can be found using Dijkstra's shortest-path algorithm~\citep{dijkstra1959note}: in each state, either the selected action is on the shortest path from this state, or it is not.
%~
%~ Note that this needs to be computed only once for each goal, independently of the number of trajectories we consider.
%~ Once these distances are pre-computed, we can use them to compute the agent mixture's probability for a given trajectory.
%~
The resulting mixture is computed with \cref{eq:magt,eq:eps_integrated}.

\subsubsection{The switching prior}\label{sec:switch-prior}
An interesting point made by \citet{baker2009action} is
that people often switch from one goal to another in the middle of a trajectory.
In order to take such behaviours into account, we will also use
\citet{veness2012context}'s switching prior technique~\citep{volf1998switching}%,koolen2013universal},
which is an efficient mixture over all sequences of models (here, all possible sequences of goals), that keeps a probability of $\frac{1}{t+1}$ of
switching at time $t$ from the current goal to a different one---and thus has a probability of $\frac{t}{t+1}$ of keeping the current goal.

Unfortunately, the switching prior does not seem to cooperate well with the integration over $\epsilon$ in \cref{eq:eps_integrated}.
Therefore, instead of using \cref{eq:eps_integrated}, we use a
%n explicitly computed
mixture of a fixed number $\Neps:=50$ of values for $\epsilon\in\{\frac{k}{\Neps-1}\}_{k \in[0..\Neps-1]}$,
which is sufficient for the purposes of this demonstration.%
\footnote{With $50$ different values, the performance of the mixture may start to degrade after a few hundreds steps, but the considered trajectories in this demonstrator are usually shorter.}

With $\allPol_{\epsilon} := \{\polufeps\}_{\uf\in\allUf}$ being the set of all policies:
\begin{align*}
&\magt(\acondotoT) :=
\sum_{\epsilon=0}^{\Neps-1} \frac{1}{\Neps}
\switch_{\allPol_\epsilon}(\acondotoT), \\
&\switch_{\allPol}(\act_t \cond \obs_{<t}, \act_{<t}) :=
\sum_{\pol\in\allPol} \wpolX{t} \pol(\act_t \cond \obs_{<t}, \act_{<t}), \\
&\wpolX{t+1} := \frac{t}{t+1}\frac{\wpolX{t}\pol(\act_t \cond \obs_{<t}, \act_{<t})}{\switch_{\allPol}(\act_t \cond \obs_{<t}, \act_{<t})}
+\frac{\wpolX{1}}{t+1}
\end{align*}
where the last line implements the switching update rule%
\footnote{This is a slight simplification
over \citep{veness2012context}
for readability that has a logarithmic loss
of $\log (t+1) + \log|\allPol|$ at each switch instead of $\log(t+1) + \log(|\allPol|-1)$.
%The base total loss of $\log (t+1)$ is the same.
}
with $\wpolX{1} := \frac{1}{|\allPol|}$.
If no switching is necessary, the cost (in the logarithmic loss) is bounded by $\log(t+1)$ at time $t$,
which is a rather small cost to pay.

Apart from the inversion of the MDP,
the computation time taken by the mixture
for a sequence of length $T$ is
$O(\Neps\Ngoals T)$, compared to $O(\Ngoals T)$ for the non-switching mixture of \cref{eq:magt}.

%~ \begin{align*}
%~ \magt(\actcondobspret) := \frac{1}{\Ngoals}\sum_{\uf\in\allUf} \pol_{\uf}(\actcondobspret)
%~ \end{align*}
%~ which can be efficiently computed in $O(|\allUf|AT)$, not considering the cost of solving the underlying MDP, which needs to be done only once independently of the number of trajectories that are evaluated.%
%~ \footnote{Ideally, the devices and the agents should share the same description space, but this would make the search space too large in general; Here we can consider that the system does receive both sets of observations, but that we, as the observer, decide to filter them depending on our current interpretation (agent or device).}

\subsection{Some trajectories}

\newcommand{\traj}{\hist_{1:T}}%\tau}
\newcommand{\mX}{\mixt_{?}}

\newcommand{\contexthits}{true} % true or false

% Command to display the 3 subfigures for each trajectory
\newcommand{\oldtrajfig}[2]{% name, caption
\begin{figure}[htbp!]
	\centering
	\begin{subfigure}[b]{.55\linewidth}
		\centering
		\includegraphics[width=0.9\linewidth]{agency-logs/#1.png}
		\caption{Trajectory.}
	\end{subfigure}
	\\
	\centering
	\begin{subfigure}[b]{\linewidth}
		\centering
		\begin{tabular}[width=\linewidth]{l | c | c | c}
		\input{agency-logs/#1-posts-v}
		\end{tabular}
		\caption{Posteriors of the device and agent mixtures. }
	\end{subfigure}
	%\\%\hfill
	%%
	%\vspace{0.1cm}
	%\begin{subfigure}[b]{\linewidth}
		%\centering
		%\begin{tabular}{l | c c c c}
			%Context & \multicolumn{4}{c}{Action} \\
			%in\_front,last\_action & up & down & left & right\\
			%\hline
			%\input{agency-logs/#1-hits}
		%\end{tabular}
		%\caption{Contexts hits.}
	%\end{subfigure}
	\caption{#2}%
	\label{fig:#1}%
\end{figure}%
}

\newcommand{\trajfigswitch}[3][]{% [3 columns], name, caption
%\NewDocumentCommand\trajfig{smm}{
\begin{figure}[htbp!]
	\centering
	\begin{subfigure}[b]{.55\linewidth}
		\centering
		\includegraphics[width=0.9\linewidth]{agency-logs/#2.png}
		\caption{Trajectory.}
	\end{subfigure}
	%\ifthenelse{#1}{\\}{}
	\begin{subfigure}[b]{.4\linewidth}
		\centering
		\begin{tabular}[width=\linewidth]{l | c | c | c}
		\input{agency-logs/#2-posts-v}
		\end{tabular}
		\begin{align*}
			v_1 &= -\ln P(\traj | \mX) \\
			v_2 &= P(\mX | \traj) \\
			v_3 &= -\ln P(\mX | \traj)
		\end{align*}
		\caption{Posteriors of the device and agent mixtures. }
	\end{subfigure}
	\ifthenelse{ \equal{#1}{} }{}{
	\\%\hfill
	\vspace{0.1cm}
	\begin{subfigure}[b]{\linewidth}
		\centering
		\begin{tabular}{l | c c c c}
			Context & \multicolumn{4}{c}{Action} \\
			in\_front,last\_action & up & down & left & right\\
			\hline
			\input{agency-logs/#2-hits}
		\end{tabular}
		\caption{Contexts hits.}
	\end{subfigure}
	}
	\caption{#3}%
	\label{fig:#2}%
\end{figure}%
}

% Context hits alone in their own figure
\newcommand{\ctxhits}[2]{
\begin{figure}[htbp!]
		\centering
		\begin{tabular}{l | c c c c}
			Context & \multicolumn{4}{c}{Action} \\
			in\_front,last\_action & up & down & left & right\\
			\hline
			\input{agency-logs/#1-hits}
		\end{tabular}
	\caption{#2}%
	\label{fig:#1-hits}%
\end{figure}%
}

% Command to display the 3 subfigures for each trajectory
\newcommand{\trajfig}[3][]{% [3 columns], name, caption
%\NewDocumentCommand\trajfig{smm}{
\begin{figure}[htbp!]
	\centering
	\begin{subfigure}[b]{.55\linewidth}
		\centering
		\includegraphics[width=0.9\linewidth]{agency-logs/#2.png}
		\caption{Trajectory.}
	\end{subfigure}
	%\ifthenelse{#1}{\\}{}
	\begin{subfigure}[b]{.4\linewidth}
		\centering
		\begin{tabular}[width=\linewidth]{l | c | c}% \@ifstar{| c}{}}
		\input{agency-logs/#2-posts-v}
		\end{tabular}
		\begin{align*}
			v_1 &= -\ln P(\traj | \mX) \\
			v_2 &= P(\mX | \traj) \\
			v_3 &= -\ln P(\mX | \traj)
		\end{align*}
		\caption{Posteriors of the device and agent mixtures. }
	\end{subfigure}
	\ifthenelse{ \equal{#1}{} \OR \equal{\contexthits}{false} }{}{
	\\%\hfill
	\vspace{0.1cm}
	\begin{subfigure}[b]{\linewidth}
		\centering
		\begin{tabular}{l | c c c c}
			Context & \multicolumn{4}{c}{Action} \\
			in\_front,last\_action & up & down & left & right\\
			\hline
			\input{agency-logs/#2-hits}
		\end{tabular}
		\caption{Contexts hits.}
	\end{subfigure}
	}
	\caption{#3}%
	\label{fig:#2}%
\end{figure}%
}

% Command to display the posteriors of the different goals.
\newcommand{\switchfig}[2]{
\begin{figure}[htbp!]
\centering
\includegraphics[width=.8\linewidth]{agency-logs/#1-switch.png}
\caption{Sequence of the posteriors of the different goals for the trajectory of \cref{fig:#1}#2.}
\label{fig:#1-switch}
\end{figure}
}

\begin{figure}[htbp!]
	\centering
	\begin{subfigure}[b]{.55\linewidth}
		\centering
		\includegraphics[width=0.9\linewidth]{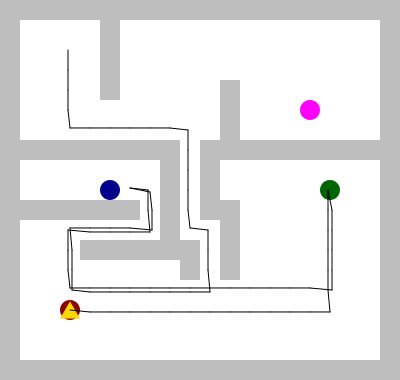}
		\caption{Trajectory.}
	\end{subfigure}
	\\
	\centering
	\begin{subfigure}[b]{\linewidth}
		\centering
		\begin{tabular}[width=\linewidth]{l | c | c | c}
		\input{agency-logs/switchB-posts-v}
		\end{tabular}
		\caption{Posteriors of the device and agent mixtures. }
	\end{subfigure}
	%\\%\hfill
	%%
	%\vspace{0.1cm}
	%\begin{subfigure}[b]{\linewidth}
		%\centering
		%\begin{tabular}{l | c c c c}
			%Context & \multicolumn{4}{c}{Action} \\
			%in\_front,last\_action & up & down & left & right\\
			%\hline
			%\input{agency-logs/#1-hits}
		%\end{tabular}
		%\caption{Contexts hits.}
	%\end{subfigure}
	\caption{Switching goals using a the switching agent model.}%
	\label{fig:switchB}%
\end{figure}

Some sample trajectories and associated results are given in \cref{fig:magenta,fig:random,fig:followwalls,fig:circle,fig:epsblue,fig:switchB,fig:switchB-switch}.%fig:random-switch}.
We report the negative log likelihood (NLL) for both device and agent mixtures,
remembering that $P(\traj|\text{agent}) = \magt(\traj)$ where we use $\traj$ as an abbreviation of $\oconditoT$.
We also report the posteriors $P(\mX|\traj)$ of the device and agent mixtures $\mX\in\{\mdev,\magt\}$ in the global mixtures $\msys$
 along with their negative log values as the latter are usually more informative, as they can be interpreted as complexities or relative losses.
The switching prior is used only for the trajectory of \cref{fig:switchB}, as for the other trajectories switching is similar to not switching.

%%%% The following pictures where generated from "Dropbox/Prog/Racket/where-are-the agents.rkt".

%Take same sequence lengths
%- A trajectory that tricks humans to think it's definitely an agent, but then we explain the very simple behaviour of the system, they change their mind.
%Related to the example of an ant walking on a beach,  of Herbert Simon 1996.

\paragraph{Running in circles.}(See \cref{fig:circle}.)
This behaviour is a prototypical example of a system behaving more like a
device than like an agent: the behaviour is very simple to explain
in terms of instantaneous reactions without referring to some goal.
\trajfig{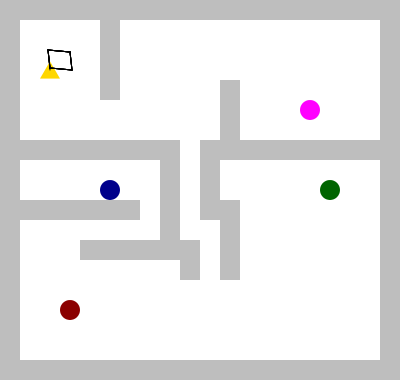}{The system is running in circles for 25 steps.}

\paragraph{Rational behaviour.}(See \cref{fig:magenta}.)
This behaviour is strongly described as that of an agent. Indeed,
it appears that it is going as fast as possible to the magenta balloon.
A device description is however still relatively simple,
as witnessed by the low relative complexity of the device mixture's posterior.
%\newcommand{\captionmagenta}{The system goes straight to the magenta balloon.}
%\input{agency-logs/magenta.tex}
%\trajfig[yes]
\trajfig{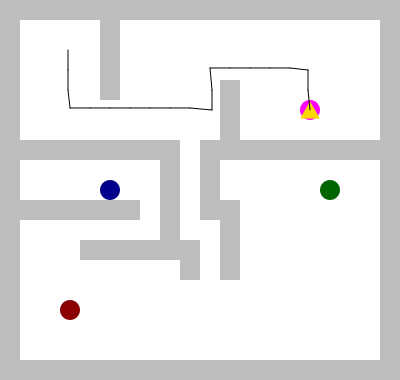}{The system goes straight to the magenta balloon.}

\paragraph{Suboptimal trajectory toward the blue balloon.}(See \cref{fig:epsblue}.)
The system attains the blue balloon after 66 steps, whereas the fastest path requires only 36 steps.
The system is still considered as an agent because of the difficulty to attain the blue balloon,
which compensates for the suboptimality of the trajectory.
\trajfig{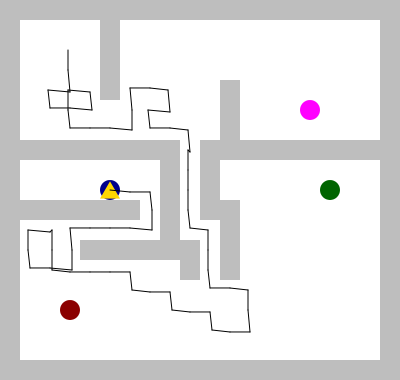}{The system is going toward the blue balloon in a suboptimal way.}

\paragraph{Following walls.}(See \cref{fig:followwalls}.)
This is another example of a behaviour that is typical of a reactive system
that acts without purposes.
This trajectory seems to be more agent-like than a random one or running in circles, and one may be tempted
to describe the behaviour of the system as ``it wants to avoid walls''.
However, when described with a simple deterministic reactive system without intentions
(``when there is a wall in front, turn right''),
it seems to lose its agency aspect.
%\comment{Although... The current goal system cannot consider walls as goals. Should we instead
%implement the goal function based on the contexts too?}
%\newcommand{\captionfollowwalls}{The system turns when facing a wall.}
%\input{agency-logs/followwalls}
\trajfig{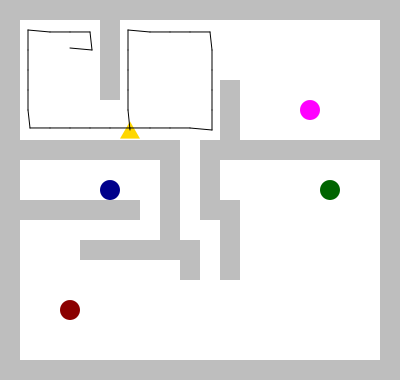}{The system turns when facing a wall.}

\paragraph{Switching goals.}(See \cref{fig:switchB}.)
The system looks like it is going first toward the magenta balloon, but before reaching it switches to going to the green balloon.
This time, for the agent's mixture we use the switching prior model described in \cref{sec:switch-prior}.
We also report the log likelihood of the trajectory for the non-switching model for information:
without the switching prior, the behaviour toward either the blue or the green balloons
is very suboptimal, and thus (without a switching prior) it is easier to consider the trajectory as generated by a device rather than an agent.
%~ \ifthenelse{\equal{\contexthits}{true}}{:
%~ as can be seen in \cref{fig:switchB-hits}, the context hits almost deterministic}{}
The posteriors of each goal along the trajectory is shown in \cref{fig:switchB-switch}.
%A few comments:
%\begin{itemize}
%~ \item The best $\epsilon$ for agents stays at 0 all the time as reported on the graph, which means
%~ that the switches are not interpreted as non-optimal behaviour.
%\item
Between steps 3 and 19, the system seems to go to any other goal than the magenta one,
and this becomes clearer starting at step 10 when the system enters the corridor.
However, the mixture cannot yet tell which goal is more likely.
%\item
Similarly, when going away from the blue balloon, the system is uncertain as to
which is the actual target now, and becomes certain it is the green balloon only after the middle corridor's entrance.
%\end{itemize}
%\newcommand{\captionswitchB}{Switching goals using a the switching agent model.}
%\input{agency-logs/switchB}
% name, caption

\switchfig{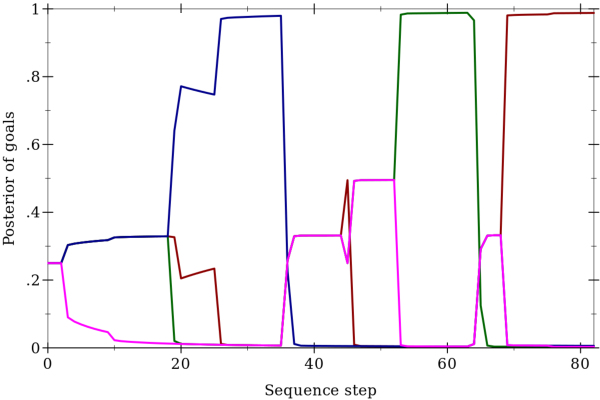}{ using a switching prior}
%\comment{We could actually make an agent mixture of the two agent mixtures instead!}

\paragraph{Random behaviour.}(See \cref{fig:random}.)
A random behaviour is difficult to explain both in terms of a device and in terms of an agent, and thus leads to a high NLL
in both cases: The context hits (see \cref{fig:random-hits}) have high entropy, and the best value for an $\epsilon$-greedy agent policy is high too
(around 0.6).
% (see \cref{fig:random-switch}).
%Both NNL are also very close to each other, which means that the global mixture is quite uncertain.
%The slight advantage for the agent mixture can be explained by the higher likelihood of the magenta goal (see \cref{fig:random-switch});
%all other goals are masked by the blue line (the sum of the posteriors is always 1).
%The graph also shows that the best value for $\epsilon$ is high,
%which is consistent with a random behaviour.
%\newcommand{\captionrandom}{The system is choosing its action uniformly randomly for 100 steps.}
%\input{agency-logs/random}
%\switchfig{random}{}
%
\trajfig[]{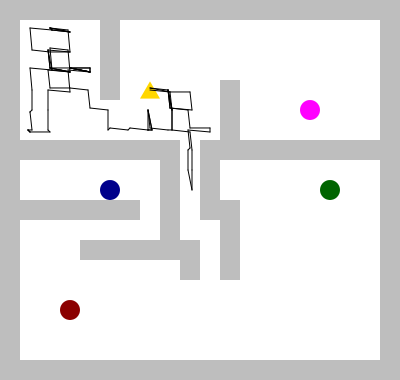}{The system is choosing its action uniformly randomly for 100 steps.}
\ifthenelse{\equal{\contexthits}{true}}{

\begin{figure}[htbp!]
		\centering
		\begin{tabular}{l | c c c c}
			Context & \multicolumn{4}{c}{Action} \\
			in\_front,last\_action & up & down & left & right\\
			\hline
			\input{agency-logs/random-hits}
		\end{tabular}
	\caption{Context hits for the random experiment (``-'' = 0).
%Dashes ``-'' mean 0. %Only contexts observed on the trajectory are displayed.
}%
	\label{fig:random-hits}%
\end{figure}%

}{}

%%%%%%%%%%%%%%%%%%%%%%%%%%%%%%%%%%%%%%%%%%%%%%%%%%%%%%%%%%%%%%%%%%%%%%%%
\vspace{-.7cm}
\section{Conclusion}
Every physical system can be described as either an agent (which pursues goals) or a device (which responds mechanically to its inputs). Hence we ask the question of subjectively \emph{how much sense} it makes to call the system an agent or a device; we quantify the answer in the form of a posterior %Bayesian
 probability. This subjective probability takes into account the observer's intrinsic biases and background knowledge.

We formalize the idea using inverse reinforcement learning techniques for agents (roughly, given a sequence of actions and observations, find the best goal and  $\epsilon$-greedy policy for this goal), and sequence prediction techniques for devices (roughly, find the best $\epsilon$-deterministic policy that fits the observed behaviour), and compare the two resulting likelihoods.

The approach was validated on a simple and clear test domain with a varied set of trajectories. While the purpose of this work is to provide a mostly non-anthropocentric formalization of a definition of agency, it would be informative to %empirically
 investigate the extent to which it matches human judgements.

From a reinforcement learning perspective, the proposed approach may also be useful to design environments that can help maximize ``agenthood'', that is, to build agents that can thrive as agents rather than performing device-like tasks.

\subsection*{Acknowledgements.}
This paper has emerged from the discussions that took place at the 2016 SAB workshop on ``Mathematical and philosophical conceptions of agency'', organized by Simon McGregor.%
\footnote{\url{http://www.sab2016.org/index.php/2-uncategorised/13-workshop1}}
Thanks also to
Peter {}Dayan,
Tom {}Erez,
Chrisantha {}Fernando,
Nando de {}Freitas,
Thore {}Graepel,
Hado Van {}Hasselt,
Andrew {}Lefrancq,
Sean {}Legassick,
Joel Z. {}Leibo,
Jan {}Leike,
R\'emi {}Munos,
Toby {}Ord,
Pedro {}Ortega and
Olivier {}Pietquin.
%% myclip | sort -t'{' -k2
%for their interesting discussions on this subject and punctual help.

%%%%%%%%%%%%%%%%%%%%%%%%%%%%%%%%%%%%%%%%%%%%%%%%%%%%%%%%%%%%%%%%%%%%%%%%
%\section*{Appendix}

\bibliographystyle{abbrnamed}%unsrtnat}
\bibliography{biblio}
\end{document}